# SimMOF: AI agent for Automated MOF Simulations

Jaewoong Lee[1a], Taeun Bae[1a], Jihan Kim[a*]

[a]Department of chemical and Biomolecular Engineering, Korea Advanced Institute of Science and Technology (KAIST), 291 Daehak-ro, Yuseong-gu, Daejeon 34141, Republic of Korea

*Correspondence to: Jihan Kim (jihankim@kaist.ac.kr)

[1]*These authors contributed equally to this work.*

**Abstract**

Metal-organic frameworks (MOFs) offer a vast design space, and as such, computational simulations play a critical role in predicting their structural and physicochemical properties. However, MOF simulations remain difficult to access because reliable analysis require expert decisions for workflow construction, parameter selection, tool interoperability, and the preparation of computational ready structures. Here, we introduce SimMOF, a large language model based multi agent framework that automates end-to-end MOF simulation workflows from natural language queries. SimMOF translates user requests into dependency aware plans, generates runnable inputs, orchestrates multiple agents to execute simulations, and summarizes results with analysis aligned to the user query. Through representative case studies, we show that SimMOF enables adaptive and cognitively autonomous workflows that reflect the iterative and decision driven behavior of human researchers and as such provides a scalable foundation for data driven MOF research.

## Introduction

Metal–organic frameworks (MOFs), composed of metal clusters and organic ligands arranged into periodic network topologies, are highly tunable porous materials that have attracted considerable attention for applications in gas storage, separation, catalysis, and electronic devices.[1-3] The nearly unlimited combinations of metal nodes, organic ligands, and network topologies endow MOFs with high design flexibility, rendering computational chemistry indispensable for exploring and understanding the vast MOF design space. Computational methods such as density functional theory (DFT), molecular dynamics (MD), and grand canonical Monte Carlo (GCMC) simulations are indispensable tools for predicting a wide range of physicochemical properties and behaviors, including structural, thermodynamic, and dynamical characteristics of materials.[4-6] Many computational simulations have also been used to identify optimal MOFs for various applications.[7-9]

However, computational chemistry demands advanced expertise, and as such, reliable MOF simulations involve a series of interdependent and nontrivial tasks that extend beyond routine calculation setup. These tasks include the construction of chemically consistent structural models and the formulation of appropriate calculation plans. Researchers must also carefully determine numerous simulation parameters, such as atomic charges, force fields, and periodic boundary conditions. In addition, meaningful interpretation of simulation results often requires further analyses, including post-processing, follow-up calculations, and the identification and resolution of simulation errors, which add another layer of complexity.[10,11] These tasks rely heavily on the empirical judgment of experienced computational chemists, and their complexity, combined with limited interoperability among simulation tools, poses a significant barrier for many of the researchers. A workflow that can simplify and automate many of these steps can be invaluable for expediting materials discovery.

Such automation must also accommodate the breadth of MOF research. Because MOFs are investigated for diverse applications, different target properties and length scales require the coordinated use of multiple simulation methods. A general-purpose platform should therefore support not only the selection and execution of appropriate computational tools, but also the construction of chemically meaningful models that account for framework–guest interactions. Moreover, because hypothetical MOFs can be generated readily, validated computational protocols should be transferable from chemically or structurally related systems to newly generated structures. The enormous size of the accessible MOF design space further necessitates efficient prescreening strategies that prioritize promising candidates before more computationally demanding calculations are performed.

Recent advances in large language models (LLMs) have demonstrated their ability to perform a wide range of complex tasks, including information retrieval through retrieval-augmented generation and decision-making based on contextual reasoning.[12-15] Trained on massive and diverse datasets, LLMs have acquired broad background knowledge spanning computational chemistry and materials science.[16,17] Beyond simple text generation, LLMs can coordinate multi step tasks, interface with external tools, and adapt their actions based on intermediate outcomes. Owing to these capabilities, LLMs have recently emerged as a promising approach for overcoming challenges in computational chemistry automation.[18,19] Beyond computational workflows, LLM-based agentic systems have also begun to demonstrate promise as experimental collaborators in chemical and materials research by assisting with experimental planning, protocol interpretation, tool use, and automated execution.[20,21] These advances suggest a broader shift toward closed-loop discovery workflows that integrate hypothesis generation, computation, experiment, and data-driven feedback.

Despite these advances, automated workflows in computational chemistry have primarily been developed in small-molecule quantum chemistry settings. For example, Zou et al.[22], Pérez-Sánchez et al.[23], and Hu et al.[24] introduced multi-agents system that translates natural-language prompts into quantum chemistry workflows. Related efforts have also begun to extend LLM-agent automation to porous-material simulations, including RASPA-based zeolite adsorption setup and force-field extraction.[25] However, MOF simulations remain substantially more challenging due to their structural heterogeneity, large system sizes, and the need to coordinate multi-step protocols across diverse tools. Existing MOF-focused automation efforts often target narrowly defined objectives (e.g., $CO_2$ adsorption or DFT structure optimization) and require nontrivial reconfiguration when extended to new tasks or material classes.[26-28] Consequently, there remains a need for flexible, scalable, and general-purpose automation that can systematically translate user intent into end-to-end MOF simulation workflows.

To remedy this, we propose SimMOF, an LLM-based multi-agent framework for automating end-to-end computational workflows in MOF research. SimMOF accepts natural language queries as input and autonomously performs workflow construction, simulation execution, and result generation. Internal reasoning is incorporated to enable self-debugging and to reduce the need for human intervention. Unlike existing LLM based approaches that focus mainly on isolated steps such as input generation or program execution, SimMOF is designed to automate the complete research process of MOF simulations. It mimics the reasoning and decision-making patterns of human experts by interpreting research objectives and contextual information. SimMOF can execute multiple tasks in parallel when appropriate and autonomously adapt calculation sequences during runtime. It also extends beyond basic result reporting by performing additional analyses and suggesting follow-up simulations when needed. This paper presents the overall architecture of SimMOF and its agent interaction mechanisms. A series of case studies with increasing complexity are then provided to demonstrate its scalability, adaptability, and ability to autonomously handle diverse computational scenarios in MOF research. By automating the simulation side of MOF research, SimMOF may also serve as a computational layer for future closed-loop experimental workflows, where simulation-derived feedback can help prioritize candidates, interpret outcomes, and guide subsequent experimental decisions.

# Results & Discussion

## SimMOF Architecture

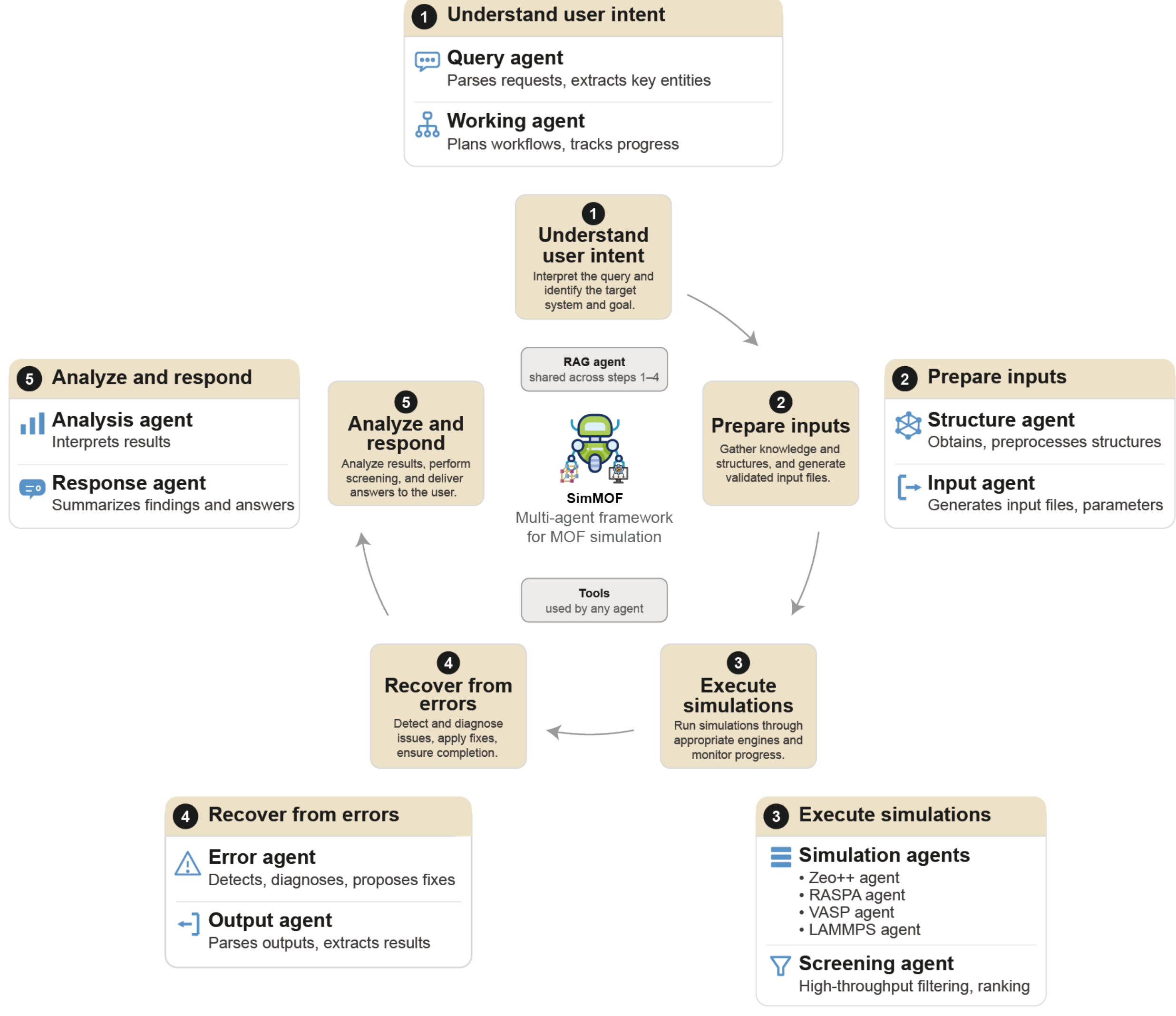

**Fig. 1** Architecture of SimMOF

SimMOF modularizes the diverse decision-making and execution steps required for MOF simulations by employing multiple LLM agents that perform specialized tasks while communicating through a shared workflow. As illustrated in Fig. 1, SimMOF operates as an iterative five-stage steps: (1) understanding user intent, (2) preparing simulation inputs, (3) executing simulations, (4) recovering from errors, and (5) analyzing and responding to the user. This representation reflects the practical workflow of computational MOF simulations, in which user requests must first be translated into executable tasks, inputs must be generated and validated, simulations must be monitored, failures must be diagnosed and corrected, and final results must be interpreted in a user-accessible form.

In the first stage, SimMOF interprets the user's natural language request and converts it into a concrete simulation objective. The query agent parses the request, extracts key entities such as the target MOF, guest molecule, property, and simulation condition, and identifies the intended computational task. Based on this information, the working agent constructs an executable workflow by decomposing the objective into one or more jobs, assigning appropriate agents, and defining dependencies between tasks. Independent jobs are scheduled for parallel execution whenever possible, whereas dependent jobs are executed in the required order. When the user request involves an analysis-oriented task, retrieval-augmented generation (RAG) is selectively invoked to retrieve relevant literature and identify how similar systems or properties have been analyzed in previous studies. This literature-informed context helps SimMOF determine appropriate analysis strategies.

In the second stage, SimMOF prepares the structural and computational inputs required for simulation. The structure agent obtains, validates, and preprocesses MOF structures from user-provided files or external structural sources. The input agent then generates the necessary simulation input files, parameters, and configuration settings according to the selected computational method. During this process, retrieval-augmented generation (RAG) is used by the input agent to retrieve relevant literature, documentation, and prior simulation protocols, allowing SimMOF to select or refine input settings based on established practices for similar systems and target properties. For example, RAG can provide contextual information for choosing simulation conditions, force-field settings, or molecule models when such choices are not fully specified by the user. This stage ensures that the workflow designed by the working agent is converted into executable input files that are consistent with the target simulation engine, user-specified conditions, and literature-informed simulation practices.

In the third stage, SimMOF executes the requested simulations using specialized simulation agents. Zeo++ agent[29] performs geometric and topological analyses, RASPA agent[30] carries out grand canonical Monte Carlo (GCMC) simulations for adsorption phenomena, VASP agent[31] conducts density functional theory (DFT) calculations for electronic structure and energetics, and LAMMPS agent[32] enables molecular dynamics simulations to evaluate dynamical, transport, and mechanical properties. Each simulation agent manages the execution process for its corresponding engine, including input validation, job submission, execution monitoring, and output collection. When the user request involves database-scale evaluation, the screening agent coordinates high-throughput filtering, candidate selection, and ranking based on the accumulated simulation results. For screening tasks, the screening agent can also use RAG during workflow construction to reference relevant literature and prior screening protocols when defining screening criteria, property thresholds, and ranking strategies. These agents are automatically invoked according to the workflow plan generated from the user's objective, allowing SimMOF to execute the required computational analyses without requiring users to manually operate individual simulation programs.

In the fourth stage, SimMOF improves robustness through automatic error recovery. The output agent parses simulation outputs and extracts relevant results, while the error agent inspects log files, detects abnormal termination or inconsistent outputs, diagnoses the likely cause of failure, and proposes corrective actions. During this process, the error agent uses retrieval-augmented generation (RAG) to retrieve relevant information for similar error cases, helping it identify failure patterns and select appropriate recovery actions. When possible, SimMOF applies these corrections iteratively by modifying input files, adjusting workflow settings, or rerunning failed jobs. This feedback mechanism allows the system to recover from common simulation errors with minimal user intervention.

In the final stage, SimMOF analyzes the completed results and communicates them to the user. The analysis agent interprets the parsed outputs in relation to the original simulation objective, including property values, rankings, and trends. The response agent then reformulates these findings into natural language and summarizes the key results. In this way, SimMOF not only performs the requested simulations but also provides an interpretable explanation of the obtained results.

SimMOF can be operated in either autonomous or interactive mode. In autonomous mode, user intervention is requested only when essential information is missing or when a user decision is required to proceed. In interactive mode, users can inspect and modify intermediate agent decisions before execution, including workflow construction, structure selection, force-field or molecule-model selection, input settings, and error-recovery actions. Further details on these user-in-the-loop control points are provided in Supplementary Note S1.

The resulting multi-agent architecture provides modularity and flexibility, with each agent performing a well-defined role while the overall workflow is coordinated through a shared, sequential process. The integration of specialized agents, shared tools, RAG-based knowledge retrieval, and automatic error recovery enables SimMOF to support reliable end-to-end MOF simulations through a natural language interface. Representative prompt templates and structured output schemas used to constrain the roles and outputs of the LLM-driven agents are provided in Supplementary Note S2. A detailed schematic of the data flow across agents is provided in Supplementary Figure S1.

## Core capabilities of SimMOF

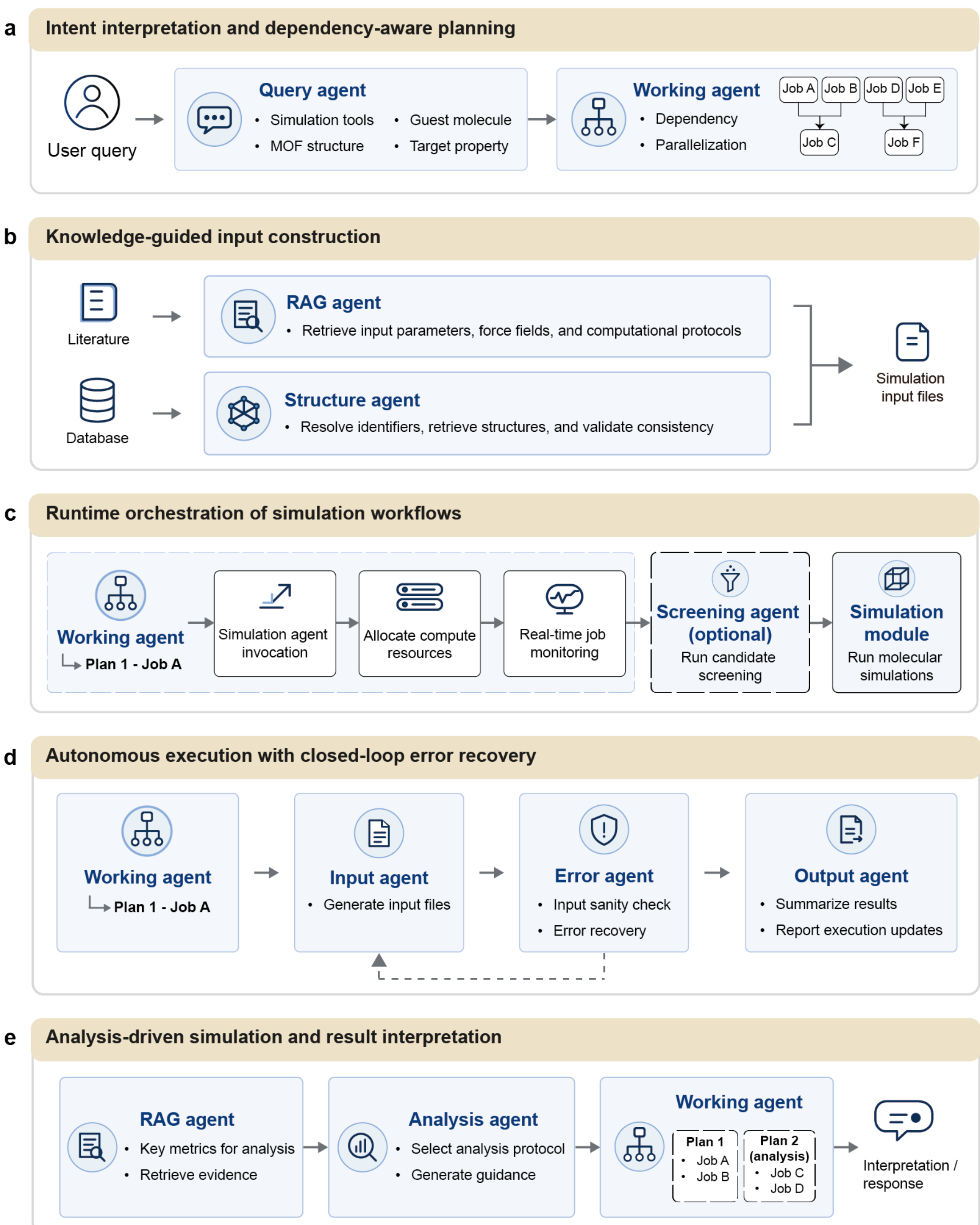

**Fig. 2** Core capabilities of SimMOF across the simulation workflow. (a) Intent interpretation and dependency-aware planning, where natural-language user queries are translated into executable job plans with explicit parallelization and dependency management. (b) Knowledge-guided input construction using RAG and database-based structure identification. (c) Runtime orchestration of simulation workflows (d) Autonomous Execution with closed-loop error recovery (e) Analysis driven simulation and result interpretation.

Fig 2 summarizes the core functions of SimMOF, including user-intent interpretation, dependency-aware workflow planning, retrieval-guided grounding of structural and literature context, and adaptive orchestration of tool execution from input construction through error recovery and result interpretation. It also delineates where LLM-based reasoning is essential across these stages.

***1) Intent Interpretation and Dependency Aware Planning***

SimMOF interprets the user's natural-language query to identify the target system and the requested property or task type. Based on the extracted intent, the working agent constructs an execution plan by decomposing the request into multiple jobs and defining their execution order. Independent jobs are scheduled for parallel execution whenever possible, where dependencies are enforced through job-level and plan-level constraints to ensure correctness. This process enables SimMOF to translate ambiguous research intents into concrete, executable workflows and to efficiently coordinate multiple simulation tools without requiring manual workflow design by the user.

***2) Knowledge Guided Input Construction***

SimMOF generates simulation input files by combining chemistry-aware LLM reasoning with prior knowledge from the literature. Through retrieval-augmented generation (RAG) over MOF paper databases, SimMOF consults reported setups for chemically similar systems, including input parameters, force-field selections, and recommended computational protocols. The detailed descriptions of the RAG-based decision framework are provided in Supplementary Note S5. The retrieval performance of this RAG step was further evaluated on representative simulation queries, and the corresponding precision and recall analyses are reported in Supplementary Table S2. To ground the workflow in the correct structural models, the structure agent resolves diverse material identifiers such as common names, synonyms, chemical formula, REFCODEs, and PORMAKE[7]-style representations, then maps them to the corresponding entries in external structure databases.

For MOFs, structural models are retrieved from the CoRE MOF[33] and CSD databases[34] and subsequently checked for chemical consistency before simulation. For small molecules, structures are obtained via the PubChem API[35], and the corresponding force-field parameters are loaded from an internal force field library for downstream simulations. For RASPA and LAMMPS workflows, this internal library is a curated collection of pre-existing framework and guest-molecule force-field and molecule-definition files. By default, the force-field agent selects and assembles appropriate files from this library rather than deriving new parameters de novo. When electrostatic interactions are required, the input agent can additionally assign framework partial charges during simulation-input construction for RASPA and LAMMPS workflows. Details of the charge-assignment procedure are provided in Supplementary Note S4. For RASPA workflows, when a required parameter set is not available in the internal library but is explicitly reported in the retrieved literature, SimMOF can use the RAG evidence to construct a custom RASPA-compatible force-field entry. To assess the contribution of this retrieval step, we compared the RAG-enabled workflow with a non-RAG prompting baseline for RASPA force-field selection and literature-derived parameter construction; the detailed ablation and experimental adsorption comparison are provided in Supplementary Table S4 and S5. This procedure does not fit new force-field parameters; instead, it converts literature-reported parameters into the input format required for the simulation. The available internal force fields and the RAG-assisted custom RASPA force-field procedure are summarized in Supplementary Table S1. Detailed procedures for automated structure lookup, user-provided CIF handling, activation, validation, and fallback behavior when a requested structure cannot be identified are provided in Supplementary Note S3. A schematic overview of the MOF structure preparation and validation workflow is also provided in Supplementary Figure S6 and S7.

***3) Runtime orchestration of simulation workflows***

SimMOF manages simulation workflows through the working agent, which coordinates task execution by invoking the simulation agents required for each plan and job, or the screening agent when candidate filtering is required before expensive calculations. Before submitting each job, the working agent delegates resource selection to a Resource Allocator module. This module receives job-specific information, including the simulation software, calculation type, and target-system size, and uses an LLM-guided reasoning step to recommend a PBS-compatible resource configuration, such as the number of nodes and CPU cores per node. This design enables different jobs to be submitted with resource settings adapted to their expected computational workload, rather than using a single fixed configuration across all calculations. During execution, the status of each job is monitored in real time so that progress can be tracked continuously and completed outputs can be passed to downstream modules for subsequent processing.

***4) Autonomous Execution with closed-loop error recovery***

SimMOF enables robust and accurate MOF simulations through a closed-loop error-recovery mechanism that minimizes manual intervention. The error agent supports this process in two complementary ways: it provides iterative feedback during input preparation and verifies job outcomes after execution to trigger recovery when necessary. The input agent first generates the initial input files, which are then validated by the error agent for basic consistency as well as common formatting and parameter issues. If problems are identified, corrective feedback is issued and the input files are regenerated iteratively. Upon job completion, the error agent inspects the log files to assess whether the run finished successfully. If a failure is detected, SimMOF applies targeted corrections, updates the relevant inputs or settings, and re-executes the job. Detailed descriptions of the error-handling mechanism are provided in Supplementary Note S8.

***5) Analysis driven simulation and result interpretation***

The analysis agent is responsible for providing scientific explanations based on analyses of the user's query and the corresponding simulation results. When the user requests analysis of simulation outcomes, the analysis agent initiates the follow-up calculations required to support the interpretation (Fig 2e). In this process, the RAG agent retrieves methodological references relevant to the requested query, which are used to guide the formulation of an analysis strategy. Based on the retrieved information, additional simulation plans for analysis are generated and executed. After completion, the analysis agent interprets the results in the context of the user's analysis request and synthesizes a mechanistic explanation.

## Case study: Zeo++ simulation with user guided clarification

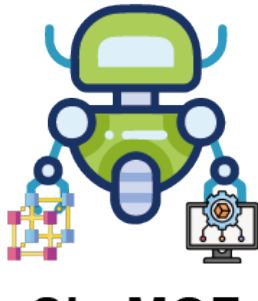

**SimMOF**

**User Query**

What is the surface area of a MOF?

**[Query agent]**

Analyzing the query and identifying the target system...
→ MOF: a MOF | Goal: Surface Area
→ Target identification: 'a MOF' is not a valid MOF identifier (generic term detected)

**[Response agent]**

Cannot compute surface area because no specific MOF was identified.

Please provide a MOF common name, cif file of MOF or REFCODE.

**User Query**

I want to calculate the surface area of UiO-66

**[Query agent]**

Analyzing query and identifying the target system...
→ MOF: UiO-66 | Goal: Surface Area

**[Working agent]**

Planned workflows:
Plan 1: UiO-66 surface area (Zeo++ Agent) | Plan 2: Final response (depends on Plan 1)

**[Response agent]**

The probe radius was not explicitly specified. The default value of 1.2 Å was applied

**[Zeo++ agent]**

| [Structure] | [Input] | [Error] | [Output] |
|---|---|---|---|
| UiO-66.cif | ./network -ha -sa 1.2 1.2 | No errors detected... | SA: 1946.02$m^2$/g |

**[Response agent]**

Converting the result into natural language...
→ ***"The UiO-66 MOF has a surface area of 1946.02 $m^2$/g."***

**Fig. 3** Zeo++ simulation workflow with interactive query refinement

Fig 3 illustrates a simple Zeo++-based workflow that highlights how SimMOF handles underspecified user queries through interactive clarification and transparent default assignment. When the user submits an ambiguous request ("What is surface area of a MOF?"), the query agent detects that the target system is not uniquely specified and determines that the request cannot be executed. The response agent then informs the user that a specific MOF identifier is required and prompts for additional information.

After the user provides a resolvable identifier (UiO-66), SimMOF proceeds to construct and execute the workflow. The working agent generates a minimal plan consisting of a Zeo++ surface-area calculation followed by response generation. When non-critical parameters are not explicitly specified (e.g., probe radius), SimMOF assigns a predefined default value and explicitly reports this selection to the user. The Zeo++ agent then runs the calculation using the retrieved structure, and the response agent returns the computed surface area in natural language.

SimMOF supports iterative user interactions during simulation to refine the workflow toward the level of detail required by the user, thereby improving reproducibility and reducing the need for manual tuning. When users cannot specify certain settings due to limited prior knowledge, SimMOF adopts commonly used default values to lower the barrier to running simulations. At the same time, SimMOF explicitly reports the required identifiers and any applied defaults to maintain traceability of the simulation setup, while allowing users to override these defaults through follow-up queries when more specific choices are desired.

## Case study: Reference guided LAMMPS simulation and error correction

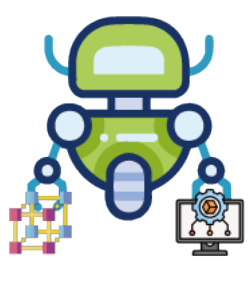

**SimMOF**

**User Query**

This is a LAMMPS input command extracted from a paper that calculates $CH_4$ diffusivity in UiO-66. I want calcualte diffusion coefficient of $CO_2$ in UiO-66.

```
pair_style      lj/cut 12.0
pair_modify   mix arithmetic
neighbor        2.0 bin
...
```

**[Query Agent]**

Analyzing query and identifying the target system...
→ MOF: UiO-66 | Guest: $CO_2$ | Goal: Diffusion Coefficient

**[Working Agent]**

Planned workflows:
Plan 1: UiO-66 diffusion coefficient (LAMMPS Agent) | Plan 2: Final response (depends on Plan 1)

**[LAMMPS Agent]**

| [Structure] | [Input] | [Error] |
| --- | --- | --- |
| UiO-66.cif<br>$CO_2$.xyz | pair_style<br>lj/cut 12.0<br>...<br>compute msd all<br>msd com yes | **Error detected!**<br>pair_style lj/cut 12.0 is not suitable for $CO_2$.<br>**Recommended configuration:**<br>pair_style lj/cut/coul/cut<br>kspace_style pppm 1.0e-5 |

**[Error Agent]**

Error found. The provided LAMMPS input was configured for $CH_4$, a nonpolar molecule, but $CO_2$ requires consideration of long-range electrostatic interactions due to its partial charges. I will modify the pair style and add a kspace solver to account for the polar molecule. Is it okay?

**[LAMMPS Agent]**

| [Structure] | [Input] | [Error] | [Output] |
| --- | --- | --- | --- |
| UiO-66.cif<br>$CO_2$.xyz | pair_style<br>lj/cut/coul/cut 12.0<br>kspace_style<br>pppm 1.0e-5<br>... | Error detecting...<br>Configuration updated.<br>Electrostatics enabled.<br>Validation passed. | D: $1.9X10^{-10} m^2/s$ |

**[Response Agent]**

Converting the result into natural language...
→ ***"Diffusion coefficient of $CO_2$ in UiO-66 is $1.9X10^{-10} m^2/s$."***

**Fig. 4** LAMMPS workflow with user-provided reference settings and input refinement.

Fig 4 presents a LAMMPS workflow that incorporates literature-derived reference settings provided by the user. The query agent extracts the analysis objective (diffusion coefficient), identifies the target system (UiO-66 with $CO_2$), and parses the provided reference settings. Based on this intent, the working agent constructs an executable plan for the LAMMPS workflow. The input agent generates a complete input deck by reusing the referenced settings as constraints and filling unspecified fields with system-consistent defaults and validation rules. In parallel, the structure agent retrieves the MOF structure file and the guest geometry file, and loads the corresponding force-field parameters from the database.

As described above, the error agent performs an additional validation step before execution by checking the simulation specification, including the generated files and system definitions, for basic consistency and common incompatibilities. In this example, the user reuses a parameter setting originally reported for $CH_4$ transport in UiO-66 and applies it to a $CO_2$ system. However, the referenced configuration (e.g., using a purely Lennard–Jones style such as "lj/cut 12.0") is not appropriate for the target system because $CO_2$ requires accounting for electrostatic interactions. The error agent detects this system-dependent mismatch, revises the input accordingly, and updates the parameter settings to a configuration compatible with the intended system (e.g., enabling Coulombic interactions and an appropriate long-range solver). The response agent then communicates the proposed modifications to the user for confirmation, ensuring that the workflow remains transparent and aligned with user intent. After user confirmation, the updated LAMMPS job is executed with the corrected settings. Upon completion, the output is parsed to extract the diffusion coefficient, and the response agent summarizes the result in natural language.

## Case Study: Parallel VASP simulations with dependency-aware scheduling

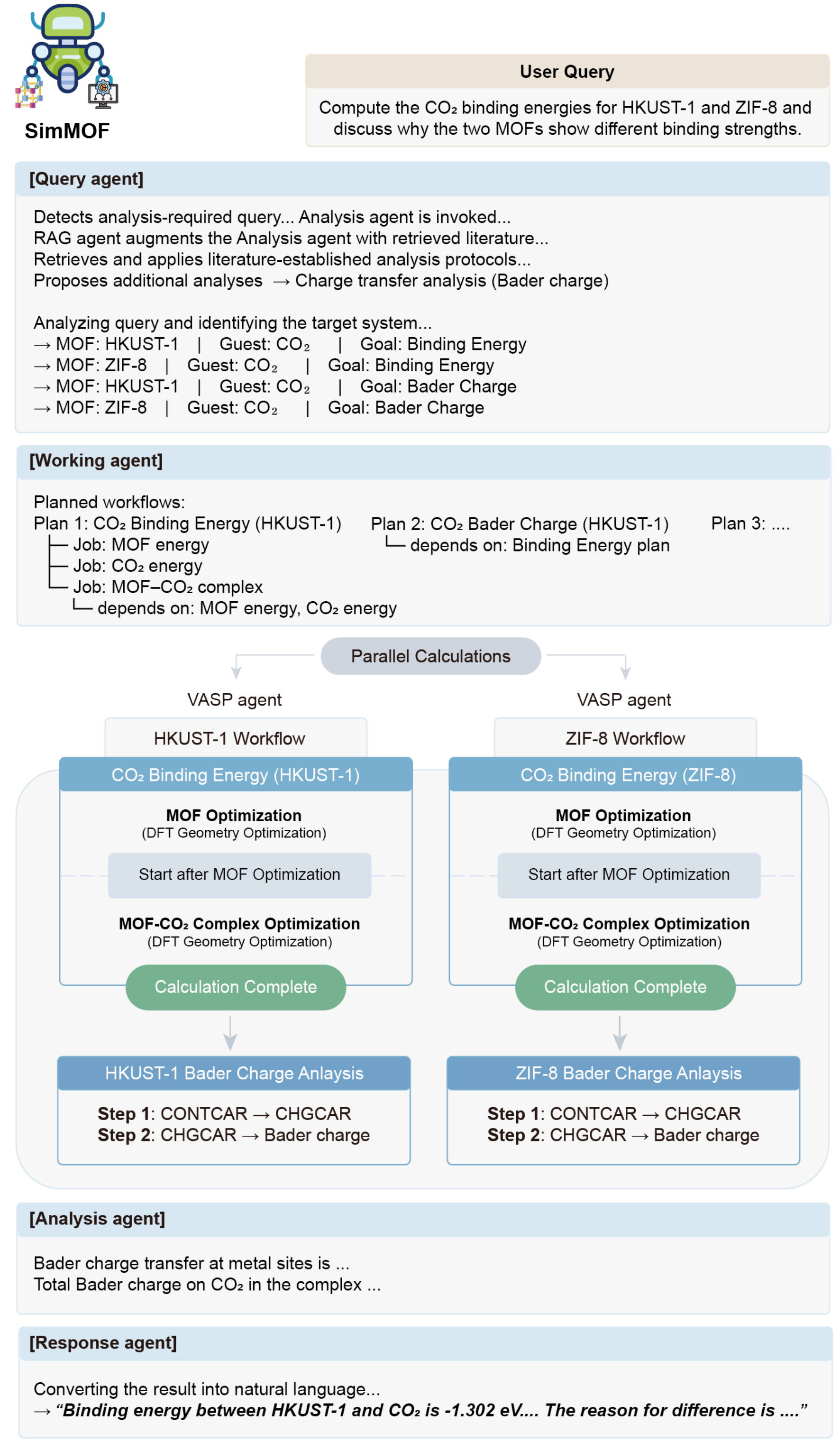

**Fig. 5** DFT workflow for comparison and analysis of $CO_2$ binding energies for two MOFs.

Fig 5 shows a comparison of $CO_2$ binding energies computed with DFT for two MOFs, HKUST-1 and ZIF-8, along with an analysis of binding energy differences. In response to the user query, the query agent identifies the request as reasoning-intensive and invokes the analysis agent accordingly. As described above, the analysis agent employs the RAG agent to retrieve methodological guidance from relevant prior studies. It then proposes additional calculations needed to rationalize differences in $CO_2$ binding energy. As a result, the working agent constructs execution plans that include both property evaluation and mechanistic analysis.

For certain simulations, such as binding energy or vibrational frequency calculations, the initial configuration can strongly affect the results. In practice, researchers generate and evaluate multiple candidate configurations by combining chemical intuition with random sampling, which in turn increases the number of calculations that must be executed and compared. At the same time, many simulation workflows impose dependencies among these calculations. For example, binding energy evaluation typically requires geometry optimization of the host framework and the guest molecule as prerequisites, followed by optimization of the host-guest complex. Accordingly, management of job scheduling is essential to execute simulation workflows efficiently.

SimMOF mirrors human expert practice by coordinating configuration exploration with dependency-aware scheduling. When multiple structural configurations are required, it generates initial candidates via random sampling and uses the MACE[36] machine-learned interatomic potential (MLIP) for low-cost prescreening. High-accuracy DFT calculations are then reserved for the selected candidates. During planning, the working agent encodes job dependencies and parallelizable steps within each plan and continuously monitors job status to launch downstream tasks as soon as prerequisites are met. This approach minimizes idle time and accelerates end-to-end workflow completion.

In the analysis workflow illustrated in Fig. 5, the following sequence is executed for both HKUST-1 and ZIF-8 while parallelizing across the two systems: host and guest structure optimization, $CO_2$ binding-energy calculations for the selected configurations, charge density generation, and Bader charge analysis using the resulting CHGCAR files. Finally, the analysis agent interprets the computed binding energies and charge-transfer descriptors to generate an explanation addressing the user's query, and the response agent returns both the quantitative results and the mechanistic interpretation.

## Case study: High-throughput screening with hierarchical filtering

SimMOF

**User Query**

I want to find MOFs with high $CH_4$ uptake in CoREMOF 2019 ASR at 10 bar

**[Query Agent]**

Analyzing query and identifying the target system...
→ Database: CoREMOF 2019 ASR | Goal: Screening MOFs good for $CH_4$ uptake at 10 bar

**[Working Agent]**

Planned workflow:
Plan 1: Database prescreening (uptake)
Plan 2: RASPA uptake simulation
└ depends on: Plan 1
Plan 3: Final response generation
└ depends on: Plan 1, Plan 2

**[Screening Agent]**

1. **Structure validation** (MOFChecker: filters out structurally invalid entries)
   – Valid entries: 9413
2. **Geometric filters** (ASE: applies an atom count constraint)
   – Matches: 9303
3. **Geometric filters** (Zeo++: evaluates pore accessibility)
   – Matches: 4331
4. **Thermodynamic proxy** (Henry constant: selected the high-performing candidates)
   – Top 1000 candidates has left

**RASPA Agents running in parallel**

**[RASPA Agent]**

HOVQAM_clean

No errors found

→ Uptake: 12.4636 mol/kg

**[RASPA Agent]**

DIWNAA_clean

No errors found

→ Uptake: 12.0731 mol/kg

**[Response Agent]**

Converting the result into natural language...
→ *"The result is as follows. Top 10 MOFs are HOVQAM_clean.cif, DIWNAA_clean.cif, ..."*

**Fig. 6** High-throughput MOF screening in SimMOF for $CH_4$ uptake on the CoREMOF 2019 ASR database

Fig 6 illustrates a high-throughput MOF screening workflow in SimMOF. When a user requests top-performing candidates from a large MOF database, the query agent parses the request to identify the target corpus and screening objective. The working agent then constructs an executable plan that incorporates a screening agent to narrow down the candidate set prior to computationally expensive simulations.

The screening agent serves as a filter layer that narrows down a large database to a tractable subset for subsequent simulation. It selects screening criteria aligned with the user's objective and executes each step by invoking low-cost simulation agents and tools. These steps are organized as a sequence of filters, where each filter is chosen to minimize computational overhead while preserving candidates that are likely to perform well for the target property.

To configure this strategy, the screening agent consults the RAG agent to retrieve screening heuristics from prior MOF screening studies and parameterizes them using task-specific thresholds (e.g., structure validation, atom counting) and low-cost descriptors (e.g., Zeo++ geometric descriptors, MLIP-based interaction energies), depending on the downstream simulation workflow. For example, it may impose stricter atom-count limits for DFT simulation workflows (e.g., $\leq$300 atoms), while allowing larger systems for GCMC workflows.

In the $CH_4$ uptake example shown in Fig 6, SimMOF searches the CoREMOF 2019 ASR for MOFs with high methane adsorption performance. Starting from the 11,660 MOFs in the initial database, the screening agent applies (i) structure validation to retain 9,413 feasible entries, followed by (ii) an atom-count constraint yielding 9,303 candidates. In this screening workflow, structures that failed MOFChecker[37] validation were excluded from the candidate set and recorded as validation failures, rather than being automatically repaired or passed to an error-recovery routine. It then applies (iii) a geometric accessibility check for $CH_4$, reducing the set to 4,331 MOFs. Finally, the screening agent computes a low-cost descriptor correlated with uptake, the Henry constant (KH), and selects the top 1,000 candidates for detailed evaluation. These candidates are then simulated in parallel by multiple RASPA agents. The response agent aggregates the outputs and returns a ranked list of top candidates consistent with the original query.

Figure 6 demonstrates that the screening agent does not simply apply a fixed set of filters, but instead constructs screening strategies based on the scientific context of the target property and system conditions. As shown in Supplementary Table S10, the screening agent selects different tools and workflows by considering factors such as the target property, dataset size, and the suitability of available proxy descriptors.

This reasoning is also reflected in the $CH_4$ uptake screening workflow. The screening agent recognizes that the KH provides a reliable proxy for uptake at low pressure, whereas its predictive relevance decreases as the pressure increases. Accordingly, it retains a relatively broad pool of 1,000 candidates at 10 bar, while using a tighter pool of 100 candidates at 0.1 bar (Supplementary Table S9).

The validity of this pressure-dependent screening strategy was further assessed in Supplementary Figures S11–S12. The results show a stronger relationship between KH and $CH_4$ uptake at 0.1 bar than at 10 bar, and a substantially smaller candidate pool is sufficient to retain high-performing materials at low pressure. These results support the physical validity of the screening agent's reasoning in adjusting screening aggressiveness according to the expected reliability of KH.

## Benchmarks

*Component level evaluations*

**Table 1** Agent-level benchmark accuracy.

| Model | Query agent | Working agent | Simulation agent handoff |
|---|---|---|---|
| GPT-5 | 25/30 | 25/30 | 26/30 |
| GPT-5.4-mini | 19/30 | 19/30 | 19/30 |
| **GPT-5.5** | **28/30** | **28/30** | **27/30** |
| llama3.1:8b[38] | 18/30 | 12/30 | 16/30 |
| qwen2.5:7b[39] | 20/30 | 20/30 | 22/30 |
| qwen2.5:14b[39] | 22/30 | 22/30 | 20/30 |
| deepseek-r1:8b[40] | 20/30 | 18/30 | 21/30 |
| deepseek-r1:14b[40] | 19/30 | 19/30 | 20/30 |

Table 1 summarizes the component-level accuracy of the query agent, working agent, and Simulation agent handoff across different LLM backends. Query agent accuracy indicates whether the user request was correctly parsed into the required simulation tasks. Working agent accuracy indicates whether the parsed tasks were converted into an appropriate workflow by considering the required simulation plans and their dependencies. Simulation agent handoff accuracy indicates whether the resulting execution plans were correctly transferred to the appropriate supported simulation agent.

Across all models, the working agent achieved the lowest aggregate accuracy, with 163 successful cases out of 240. This lower performance likely reflects the greater difficulty of converting parsed simulation tasks into an appropriate workflow while considering the required calculations, execution order, and dependencies among tasks.

Among the evaluated models, GPT-5.5 achieved the highest overall performance, with accuracies of 28/30 for both the query agent and working agent and 27/30 for simulation agent handoff. Overall, the GPT-series models outperformed the open-source models. This difference may reflect both the intrinsic capabilities of the models and the fact that the prompt-engineering process was developed and optimized primarily using GPT-5.5 as the base model. The detailed queries used in the evaluation and representative failure cases are provided in Supplementary Tables S12, S16 and S17.

*Simulation Input Generation*

**Table 2** Evaluation of simulation input generation across different LLM backends with and without RAG

| Model | Zeo++ | | RASPA | | LAMMPS | | VASP | |
|---|---|---|---|---|---|---|---|---|
| | w/o RAG | with RAG | w/o RAG | with RAG | w/o RAG | with RAG | w/o RAG | with RAG |
| gpt-5.4-mini | 0/10 | 9/10 | 9/20 | 15/20 | 5/20 | 12/20 | 14/20 | 16/20 |
| gpt-5 | 2/10 | **10/10** | 10/20 | 16/20 | 8/20 | 10/20 | 15/20 | 17/20 |
| gpt-5.5 | **6/10** | **10/10** | **16/20** | **18/20** | **15/20** | **18/20** | **16/20** | **18/20** |
| deepseek-r1:14b | 1/10 | 7/10 | 10/20 | 11/20 | 12/20 | 13/20 | 3/20 | 15/20 |
| deepseek-r1:8b | 1/10 | 7/10 | 1/20 | 8/20 | 7/20 | 12/20 | 1/20 | 10/20 |
| llama3.1:8b | 2/10 | 7/10 | 7/20 | 8/20 | 7/20 | 11/20 | 2/20 | 8/20 |
| qwen2.5:14b | 2/10 | 7/10 | 8/20 | 14/20 | **15/20** | 16/20 | 4/20 | 15/20 |
| qwen2.5:7b | 1/10 | 6/10 | 7/20 | 10/20 | 12/20 | 13/20 | 1/20 | 12/20 |

To evaluate the dependence of simulation input generation on the choice of LLM backbone and the contribution of simulator-specific retrieval, we tested multiple proprietary and open-source LLMs using the same structured prompts under conditions with and without RAG. As summarized in Table 2, the use of RAG improved the success rate for all evaluated models and simulation packages, although the magnitude of improvement varied across models and simulators.

Among the evaluated models, GPT-5.5 achieved the highest overall performance, with the number of successful cases increasing from 53 out of 70 without RAG to 64 out of 70 with RAG. Smaller open-source models generally showed lower baseline performance and a greater dependence on retrieved evidence. For example, the performance of DeepSeek-R1 8B increased from 10 successful cases out of 70 without RAG to 37 out of 70 with RAG. Substantial improvements were also observed for Qwen2.5 14B and GPT-5.4-mini. These results indicate that simulator-specific RAG evidence can partially compensate for the limited internal knowledge of smaller LLM backbones. However, RAG did not completely eliminate the performance differences among the backbone models. This indicates that successful simulation input generation depends on both the reasoning capabilities of the underlying LLM and access to relevant external evidence.

Although performance improvements with RAG were observed across all four simulation packages, the magnitude of the RAG effect varied among the packages. The largest improvement was observed for Zeo++. We infer that this may be attributed to the relatively simple and well-defined command structure of Zeo++, which enables relevant evidence to be retrieved effectively and incorporated directly into command generation. The evaluation queries and corresponding failure cases are provided in Supplementary Note S6.

*Reproduction of Selected Simulation Results*

**Table 3** Comparison of SimMOF generated simulation results with reference values for representative MOF properties calculated using Zeo++, RASPA, LAMMPS, and VASP.

| Simulation Tool | MOF | Property | Unit | Reference Value | SimMOF Value | Error(%) | Ref |
|---|---|---|---|---|---|---|---|
| Zeo++ | PUPJER | LCD | Å | 11.35 | 11.35 | 0.00 | [41] |
| Zeo++ | RUPTED | LCD | Å | 4.59 | 4.59 | 0.00 | [41] |
| Zeo++ | NORPIV | PLD | Å | 3.55 | 3.55 | 0.00 | [41] |
| Zeo++ | NUYQUU | PLD | Å | 3.47 | 3.47 | 0.00 | [41] |
| RASPA | CICYIX | $N_2$ uptake 77K, 200Pa | $cm^3/g$ | 75.38 | 75.57 | 0.26 | [42] |
| RASPA | FIGXEY | $N_2$ uptake 77K, 200Pa | $cm^3/g$ | 150.61 | 150.62 | 0.01 | [42] |
| RASPA | tobmof-7165 | $CH_4$ uptake 298K, 65bar | $cm^3/g$ | 184.00 | 184.56 | 0.30 | [42] |
| RASPA | tobmof-7187 | $CH_4$ uptake 298K, 65bar | $cm^3/g$ | 233.00 | 233.85 | 0.37 | [42] |
| RASPA | tobmof-7165 | $H_2$ uptake 243K, 100bar | g/L | 9.50 | 9.13 | 3.91 | [42] |
| RASPA | tobmof-7187 | $H_2$ uptake 243K, 100bar | g/L | 11.20 | 10.52 | 6.11 | [42] |
| LAMMPS | BUKRUW01 | $O_2$ diffusivity | $cm^2/g$ | 2.59.E-04 | 2.62.E-04 | 1.06 | [43] |

| LAMMPS | MAPCIP | $O_2$ diffusivity | $cm^2/g$ | 2.89.E-04 | 2.97.E-04 | 2.60 | [43] |
|---|---|---|---|---|---|---|---|
| VASP | GUXQAR | bandgap | eV | 0.08 | 0.07 | 7.50 | [44] |
| VASP | RURPAW | bandgap | eV | 1.11 | 1.14 | 2.49 | [44] |
| VASP | NUYQUU | bandgap | eV | 2.90 | 2.90 | 0.10 | [44] |
| VASP | GIFKEL | bandgap | eV | 3.18 | 3.19 | 0.07 | [44] |
| VASP | GAYGAQ | $CO_2$ binding energy | eV | -0.33 | -0.32 | 2.90 | [45] |
| VASP | GAYGAQ | $H_2O$ binding energy | eV | -0.28 | -0.33 | 17.85 | [45] |
| VASP | GIFKEL | $CO_2$ binding energy | eV | -0.20 | -0.20 | 0.39 | [45] |
| VASP | GIFKEL | $H_2O$ binding energy | eV | -0.58 | -0.48 | 17.02 | [45] |

We further evaluated the quantitative accuracy of SimMOF by comparing simulation results with reference values across representative MOF properties, as summarized in Table 3. Representative examples of the generated working plans and simulation input files for the Table 3 benchmark cases are provided in Supplementary Figures S4 and S5, respectively. The benchmark covers geometric descriptors, adsorption properties, transport properties, and electronic properties calculated using Zeo++, RASPA, LAMMPS, and VASP. In most cases, SimMOF reproduced reference values with high accuracy, yielding small errors across a wide range of properties. Although a relatively large deviation of approximately 17% was observed for the $H_2O$ binding energy in VASP calculations, the absolute difference corresponds to only 0.05 eV, which falls within the typical error range of density functional theory calculations. The deviation in binding energy can be attributed to the generation of initial MOF-guest configurations. In the current implementation, SimMOF generates a limited number of candidate structures using random placement and selects the most stable configuration based on a machine-learned interatomic potential prior to DFT calculation. Increasing the number of sampled configurations is expected to further improve agreement with reference values. Overall, these results demonstrate that SimMOF can reliably reproduce simulation outcomes across diverse computational methods while maintaining practical accuracy for MOF research applications. However, it should be noted that the successful cases were limited to studies in which the Methods section reported the parameter settings that could affect the simulation results in sufficient detail and the structures could be used directly without expert-level manual modification. Therefore, these results should not be interpreted as evidence that SimMOF can reliably reproduce all reported simulation studies.

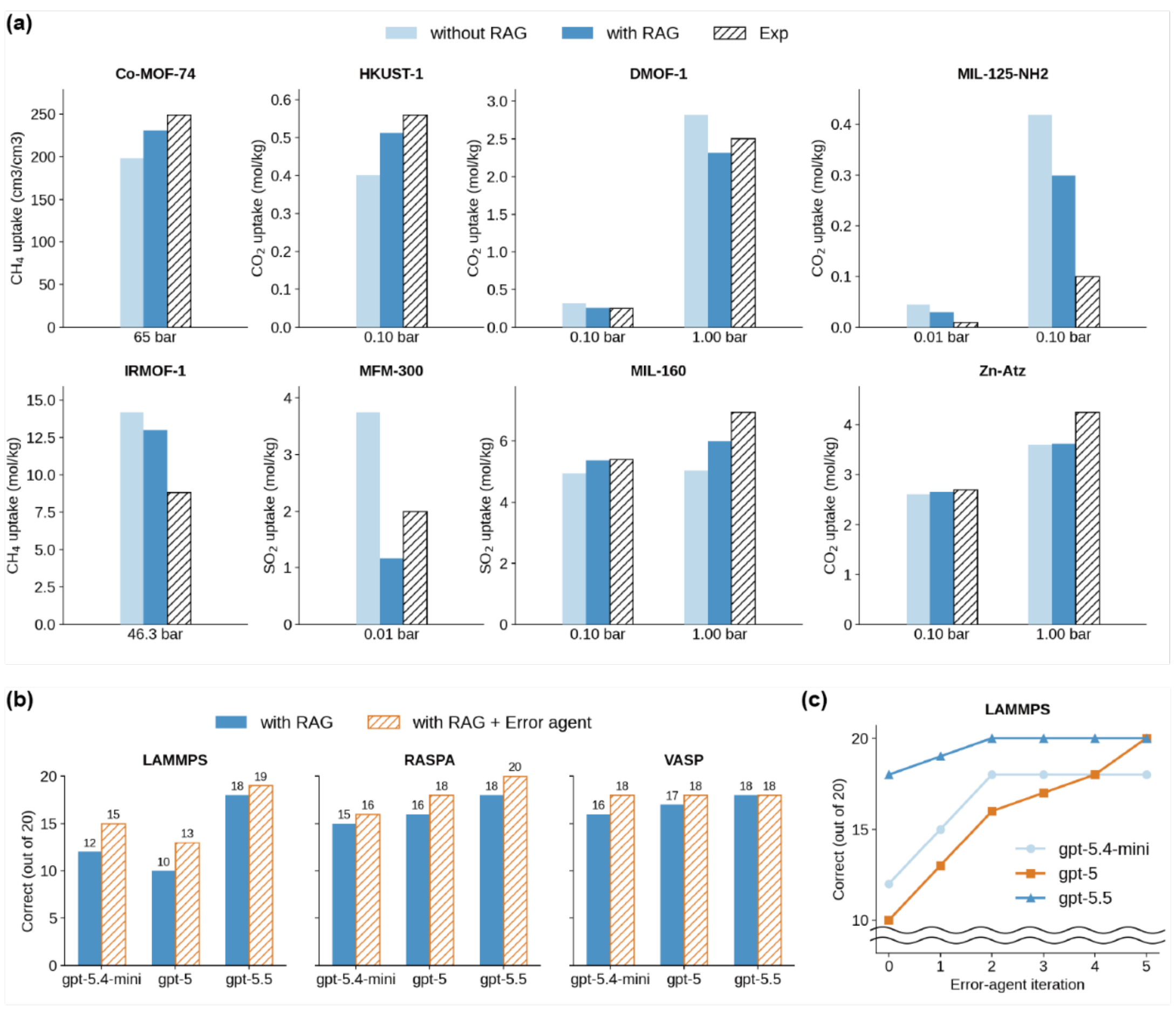

**Fig. 7** Quantitative evaluation of RAG-guided input generation and error recovery. (a) Comparison of RAG-guided force-field and charge-assignment selections with experimental values.(b) Accuracy of simulation inputs after initial generation by the input agent and subsequent correction by the error agent.(c) Accuracy of LAMMPS simulation inputs as a function of the number of correction iterations.

*RAG-Guided Parameter Evaluation*

Figure 7(a) compares the adsorption uptakes calculated using the force-field and charge-assignment settings selected by the RASPA workflows with and without RAG against the corresponding experimental values. The selected settings and calculated results are summarized in Supplementary Tables S4 and S5. Without RAG, the workflow predominantly relied on generic parameter combinations, such as UFF4MOF[46] with EPM2[47] or TraPPE[48] guest models and PACMAN[49] charges. By contrast, the RAG-enabled workflow selected more system-specific settings, including custom or mixed framework force fields, alternative guest models, and DDEC-derived charges. These selections reduced the deviation from experiment in all 12 evaluated cases, although the magnitude of improvement varied across the systems. The residual discrepancies observed for MIL-125-$NH_2$ and IRMOF-1 indicate that literature-guided parameter selection can reduce reliance on generic defaults but is not, by itself, sufficient to guarantee quantitative agreement with experimental adsorption data.

*Simulation Input Correction by the Error agent*

Figure 7(b) evaluates whether the error agent can further improve simulation inputs that still contain errors after RAG-assisted generation by the input agent. Across LAMMPS, RASPA, and VASP, the number of correct inputs generally increased after application of the error agent, with particularly clear improvements in cases showing lower initial accuracy. These results demonstrate that the error agent can interpret runtime error messages generated by the simulation programs and revise the existing inputs using evidence retrieved from the corresponding software manuals. In particular, Figure 7(c) shows that the accuracy of the LAMMPS simulation inputs increased over successive correction iterations. This finding indicates that runtime error messages provide additional diagnostic evidence that can complement incomplete retrieval results and limitations in the LLM's initial reasoning.

## Discussion and future work

Despite its demonstrated ability to emulate key aspects of expert reasoning in MOF simulation workflows, SimMOF remains bounded by the generalization capability of the underlying large language models. While the framework performs robustly for common simulation tasks and frequently encountered error patterns, its performance may degrade when facing rare failure modes or highly specialized scenarios that are underrepresented in training data. In such cases, domain-specific fine-tuning of the language model or the incorporation of structured expert knowledge may be necessary to ensure reliable decision-making and robust error recovery.

In addition, while SimMOF currently integrates a diverse set of simulation tools widely used in MOF research, its coverage of computational methods is not exhaustive. Certain simulation paradigms and analysis tasks remain outside the scope of the current framework, including advanced free-energy calculations, ab initio molecular dynamics, and direct integration with experimental feedback. Expanding the supported toolset and enabling seamless incorporation of additional simulation engines and data sources will be essential for broadening the applicability of SimMOF across diverse MOF research scenarios.

Finally, SimMOF does not yet constitute a fully autonomous, closed-loop materials discovery system. Although the framework can independently execute and adapt well-defined computational tasks such as property evaluation, screening, and workflow-level error correction, higher-level research objectives, including autonomous inverse design, and goal-driven optimization across iterative simulation cycles, still require explicit user-defined goals and supervision. Extending SimMOF toward fully closed-loop simulation and design remains a key direction for future development.

## Conclusion

In this work, we presented SimMOF, an LLM-based multi-agent framework that translates natural-language requests into executable MOF simulation workflows and enables end-to-end automation. Across case studies ranging from simple property calculations to large-scale screening, SimMOF adapts workflows to diverse objectives and computational constraints by dynamically selecting tools, and efficiently executing parallel simulations, and thereby reduces human intervention. Overall, SimMOF lowers the barrier to computational MOF research while maintaining flexibility and reproducibility for advanced tasks, providing a scalable foundation for data-driven materials discovery and future self-driving laboratory environments that can be extended to other complex materials systems and multi-step computational workflows.

## Methods

### *Large Language Model Configuration*

All agent reasoning, planning, simulation orchestration, and workflow generation were performed using OpenAI's GPT-5.5 reasoning model (API model identifier: gpt-5.5-2026-04-23) through the OpenAI Chat Completions API (https://api.openai.com/v1/chat/completions). This model is publicly available through the OpenAI API. The same model configuration was used for all LLM-driven agents in SimMOF. Additional LLMs were evaluated only for comparative benchmark analyses. The model identifiers, model sizes, and access configurations of these benchmark models are summarized in Supplementary Table S14. GPT-5.5 supports configurable reasoning effort rather than conventional sampling controls such as temperature or top-p. In this study, the default reasoning effort was used for all inference calls. No additional inference parameters were explicitly specified, and the same date-stamped model snapshot and API configuration were used throughout the study.

No fine-tuning was performed on the base language model. Instead, model behavior was controlled exclusively through prompt engineering and structured output constraints. Task-specific system prompts were carefully designed to enforce role consistency, ensuring that agents maintained well-defined responsibilities such as planning, tool routing, or screening workflow design. Structured JSON schema constraints were imposed to guarantee machine-readable outputs for all workflow definitions, simulation configurations, and decision modules. In addition, explicit hard rules were embedded within prompts to prevent undesired tool selection, numerical condition hallucination, and speculative reasoning beyond the available computational context. Deterministic formatting instructions were further incorporated to eliminate ambiguity during downstream parsing and execution.

This prompt-engineering-based control architecture ensured domain-aligned reasoning and strict procedural consistency without modifying the pretrained model weights.

***Retrieval-Augmented Generation (RAG)***

To ground agent decisions in prior literature and reduce hallucinations, we implemented a retrieval-augmented generation (RAG) module that retrieves relevant scientific passages from a local corpus and injects them as evidence into downstream agent prompts. The RAG corpus was constructed from two sources: (1) full-text publisher XML (and HTML where applicable), converted into a structured text representation during parsing, and (2) supplementary information (SI) documents, converted from PDF and DOCX formats into plain text.

***Corpus-specific chunking***

For the literature corpus, each document was split into chunks based on section boundaries, which were identified from bracketed headers (e.g., [Section Title]) inserted during parsing. Within each section, text was broken into sentences and packed into chunks up to 700 characters (max_chars = 700). Chunks shorter than 200 characters (min_chars = 200) were merged with the next segment to avoid overly small fragments. One sentence of overlap was kept between consecutive chunks to reduce information loss at boundaries. The References section was excluded from chunking.

For the SI corpus, documents did not contain section headers, so a different chunking approach was used. Tables were identified and kept as single chunks where possible; if a table was too long, it was split by rows while repeating the header row in each piece so that column labels were always preserved. Non-tabular text was chunked using the same sentence-packing logic as the full-text corpus (max_chars = 700, min_chars = 200), without sentence-level overlap.

***Embedding model and vector index***

Chunks were embedded using a SentenceTransformer encoder (default: sentence-transformers/all-MiniLM-L6-v2)[50]. Embeddings were L2-normalized (normalize_embeddings=True) and indexed using FAISS[51] with an inner-product index (IndexFlatIP). Under normalization, inner-product scoring is equivalent to cosine similarity, providing stable similarity search behavior. For each embedded chunk, we stored provenance metadata including the source filename and chunk identifier, enabling traceable evidence attribution during downstream agent reasoning.

***Batched indexing for scalability***

To efficiently process large corpora, both file I/O and embedding were performed in batches. Documents were read in file batches (e.g., file_batch_size=200), while chunks were embedded and added to the index in embedding batches (e.g., encode_batch_size=512). This approach controlled memory usage while maintaining high throughput. The resulting FAISS index and metadata were serialized to disk (index.faiss and metadata.pkl) under a model-specific directory to support reproducible retrieval runs.

***Code availability***

The code and documentation for SimMOF, including resources relevant to the workflows described in this study, are publicly available at the following GitHub repository: https://github.com/skyljw0714/SimMOF